\documentclass[10pt,twocolumn,letterpaper]{article}

\usepackage{cvpr}
\usepackage{times}
\usepackage{epsfig}
\usepackage{graphicx}
\usepackage{amsmath}
\usepackage{amssymb}
\usepackage{algorithm}
\usepackage{algorithmic}
\usepackage{caption}
\usepackage{authblk}
\makeatletter
\renewcommand\AB@affilsepx{~ \protect\Affilfont}
\makeatother
\usepackage[pagebackref=true,breaklinks=true,letterpaper=true,colorlinks,bookmarks=false]{hyperref}

\cvprfinalcopy 

\def\cvprPaperID{1538} 

\DeclareMathOperator*{\argmin}{\arg\!\min}

\begin{document}

\title{Occlusion-Aware Video Deblurring with a New Layered Blur Model\vspace{-1em}}

\author[1,*]{Byeongjoo Ahn}
\author[2,*]{Tae Hyun Kim}
\author[3,$\dagger$]{Wonsik Kim}
\author[3]{Kyoung Mu Lee}
\affil[1]{Korea Institute of Science and Technology\protect\\}
\affil[2]{Max Planck Institute for Intelligent Systems}
\affil[3]{Seoul National University}

\twocolumn[{%
    \renewcommand\twocolumn[1][]{#1}%
    \maketitle
    \begin{center}
        \centering
        \includegraphics[width=\textwidth]{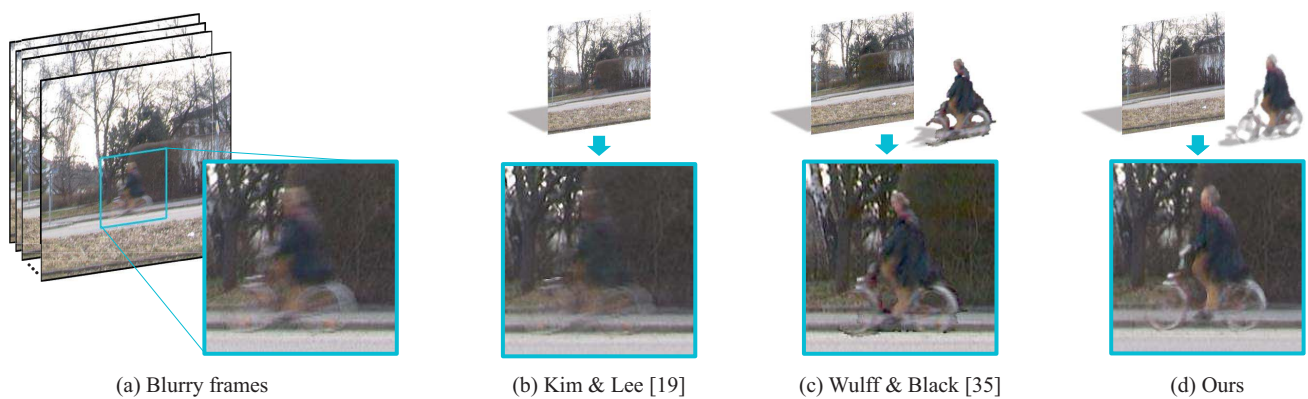}
        \captionof{figure}{Deblurring results for a scene with an occluding object. Our method produces better results at occlusion boundaries than generalized video deblurring method~\cite{kim2015generalized} and the previous layered deblurring method~\cite{wulff2014modeling}.}
        \label{fig_teaser}
    \end{center}%
}]
{
  \renewcommand{\thefootnote}%
    {\fnsymbol{footnote}}
  \footnotetext[1]{Part of this work was done while the authors were at Seoul National University.}
  \footnotetext[2]{Currently at Samsung Electronics.}
}

\maketitle
\thispagestyle{empty}

\begin{abstract}
\vspace{-0.5em}
We present a deblurring method for scenes with occluding objects using a carefully designed layered blur model.
Layered blur model is frequently used in the motion deblurring problem to handle locally varying blurs, which is caused by object motions or depth variations in a scene.
However, conventional models have a limitation in representing the layer interactions occurring at occlusion boundaries.
In this paper, we address this limitation in both theoretical and experimental ways, and propose a new layered blur model reflecting actual blur generation process.
Based on this model, we develop an occlusion-aware deblurring method that can estimate not only the clear foreground and background, but also the object motion more accurately.
We also provide a novel analysis on the blur kernel at object boundaries, which shows the distinctive characteristics of the blur kernel that cannot be captured by conventional blur models.
Experimental results on synthetic and real blurred videos demonstrate that the proposed method yields superior results, especially at object boundaries.
\end{abstract}


\section{Introduction}
\begin{figure*}
  \centering
  \includegraphics[width=\linewidth]{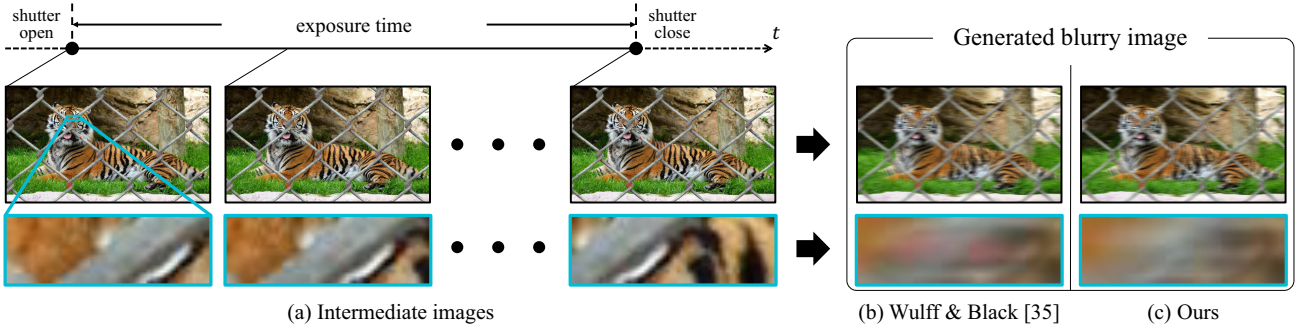}
  \caption{Counterexample that shows the difference between the actual blur generation process and the conventional layered blur model~\cite{wulff2014modeling}. Even though the tiger's left eye (red color) in the background is occluded by the the foreground fence in the entire time window between the opening and closing of the camera shutter, it is exposed in the conventional generative model~\cite{wulff2014modeling}. This error can cause severe ringing artifacts in deblurring results. Notice that our generative model successfully reflects the actual blur generation process.}
  \label{fig_counter}
\end{figure*}
\vspace{-0.5em}
Recently, deblurring technique has made a lot of progress. Early deblurring methods only focused on the blur caused by camera shake in constant depth images~\cite{cho2009fast,fergus2006removing,whyte2012non,xu2013unnatural}. Recently, however, there are some methods to handle the blurred images with depth variations~\cite{hu2014joint,lee2013dense,paramanand2013non,xu2012depth}, and there are even methods to handle object motion blur~\cite{kim2013dynamic,kim2015generalized,pan2016soft}. Object motion deblurring problem is very challenging since it requires the estimation of independent spatially-varying blur kernels.

What makes the object motion deblurring problem more difficult is the occlusions generated by intra-frame motions. While occlusions generated by \emph{inter}-frame motions cause a photo-inconsistency, occlusions generated by \emph{intra}-frame motions cause a mixture of foreground pixels and background pixels at occlusion boundaries in the blurred image.
These ambiguous pixels can lead to severe ringing artifacts in the deblurring results.

To address this problem, several methods explicitly modeled the occlusions using layered blur model~\cite{bar2007variational,chakrabarti2010analyzing,dai2009removing}. Especially, Wulff and Black~\cite{wulff2014modeling} proposed a layered blur model for the case where both layers are blurred, and obtained convincing results. They modeled a blur image as a composition of individually blurred foreground and background, and this generative model could express the layer interaction caused by occlusions. However, their layered blur model does not reflect the actual blur generation process. A blurred image is the integration of intermediate images that the camera sees while the shutter is open, which differs from their layered blur model. Figure~\ref{fig_counter} shows a counterexample. The foreground and background are moving at similar speeds in the image, and the blurred image is observed as an integration of the intermediate images. Since the tiger's left eye (red color) in the background is occluded by the foreground fence in the entire time window between the opening and closing of the camera shutter, it should not be exposed in the blurry image but it is visible in their generative model. This error can cause severe ringing artifacts in deblurring results.

In this paper, we propose a new layered blur model reflecting the actual blur generation process, and occlusion-aware video deblurring method accordingly. We enhanced the model by changing the order of a composition and a blurring of layers, so that it follows the actual blur generation process. By using the carefully designed likelihood from our layered blur model, clear foreground and background can be successfully recovered from blurred images with occluding objects. Specifically, given a set of motion-blurred video frames, our method estimates clear foreground, clear background, clear alpha blending mask, and the motion for each layer. Figure~\ref{fig_teaser} shows that our result is better than that of~\cite{wulff2014modeling}, especially at occlusion boundaries. Also, we analyze the layered blur model theoretically and experimentally. We show that the model of~\cite{wulff2014modeling} is a good approximation that is identical to our model for some specific physical situations, and also present that the blur kernels at boundaries have distinct characteristics that cannot be captured by conventional blur models.

\section{Related Work}
Early works on deblurring focused on the blur caused by camera shake in constant depth images. They are roughly categorized into spatially-invariant and spatially-varying configurations. Spatially-invariant deblurring achieved some success in single-image deblurring~\cite{cho2009fast,fergus2006removing,shan2008high,xu2010two} and video deblurring~\cite{cai2009blind}. However, the spatially-invariant blur model cannot deal with a rotational camera motion, which is a significant and common component in practical scenarios~\cite{levin2009understanding}. To overcome this limitation, some researchers parameterized the blur as a possible camera motions in a 3D space, and this approach is applied to single-image~\cite{gupta2010single,hirsch2011fast,whyte2012non,xu2013unnatural} and video deblurring~\cite{cho2012registration,li2010generating}. Although these methods solve spatially-varying motion blur in some extent, they are limited to camera shake in a constant depth and cannot handle more general depth variation or object motion problem.

In the case of blurred images including depth variations, the blur cannot be represented by a simple homography. Some methods solved this problem by casting a blur kernel estimation problem as a scene depth estimation problem~\cite{hu2014joint,lee2013dense,paramanand2013non,xu2012depth}.
These methods extended the applicability of deblurring methods. However, they are limited to static scenes, and do not take the mixture of pixels at occlusion boundaries into account.

Recently, several object motion deblurring methods have been developed. Some of the methods divided the image into segments to restore each of them independently. They divided the image using hybrid cameras~\cite{ben2003motion,tai2010correction}, based on similar motions~\cite{cho2007removing,kim2013dynamic,levin2006blind}, or under the guidance of the matting~\cite{pan2016soft}. There are also some methods without segmentations. Cho~\etal~\cite{cho2012video} used patch-based synthesis for deblurring by detecting and interpolating proper sharp patches at nearby frame. Kim and Lee approximated the blur kernel as the pixel-wise 2D linear motion and performed deblurring of dynamic scene in a single-image~\cite{kim2014segmentation} and a video~\cite{kim2015generalized}. These object motion deblurring methods perform well, but do not consider the interaction between the object and the background at occlusion boundaries.

At occlusion boundaries, blurred pixels consist of a mixture of foreground pixels and background pixels and it plays an important role for object motion deblurring. To address this problem, some authors used layered models~\cite{bar2007variational,chakrabarti2010analyzing,dai2009removing}. However, these methods assumed the background to be static and modeled the foreground motion only. Sellent~\etal~\cite{sellent2016stereo} used outlier rejections~\cite{chen2008robust} to handle the occlusions. Takeda and Milanfar~\cite{takeda2011removing} proposed a method that can deal with occlusions using a spatiotemporal approach, but it requires priorly given blur kernels and depends on time interpolators.

To deal with the general case where both the foreground and the background are moving independently, Wulff and Black~\cite{wulff2014modeling} proposed a layered blur model that consists of a composition of individually blurred foreground and background. This model included the interaction between layers and improved the performance at occlusion boundaries. However, as shown in Figure~\ref{fig_counter}, this generative model is different with the actual blur generation process and not always valid.


\section{Analysis of Layered Blur Model}
In this section, we briefly review the previous layered blur model~\cite{wulff2014modeling}, propose our new layered blur model, and compare them. The differences in these models will be proved to be greatly attributable to removing serious artifacts at occlusion boundaries in the later section.

First of all, we set a layered model for a clear image $I \in \mathbb{R}^n$ ($n$ is the number of pixels) to be:
\begin{equation}
I = (\mathbf{1} - A) \odot L_1 + A \odot L_0,
\label{eq_layer_clear}
\end{equation}
where $L_1 \in \mathbb{R}^n$ and $L_0 \in \mathbb{R}^n$ are the clear foreground and background layer, respectively, $A \in \mathbb{R}^n$ is an alpha blending mask, $\mathbf{1} \in \mathbb{R}^n$ is a vector with all components equal 1, and $\odot$ denotes element-wise multiplication. Notice that $A$ multiplied the background image ($A=1$ means a background pixel) for notational simplicity later on, inspired by~\cite{xue2015computational}.

Our goal is to express blurred images using each layer of the reference image (\ie $\{L_1, L_0, A\}$). Before expressing blurred images, we express a warped image using $\{L_1, L_0, A\}$. We assume that the appearance and shape of each layer is constant, which is a common assumption in the deblurring literature~\cite{bar2007variational,dai2009removing,wulff2014modeling}. If we let $\theta_l^i$ denotes a motion parameter for layer $l \in \{0, 1\}$ from the reference frame to the frame $i$, then the warped image $I^i$ at frame $i$ is as follows:
\begin{equation}
I^i = (\mathbf{1} - \mathbf{W}(\theta_1^i) A) \odot \mathbf{W}(\theta_1^i) L_1 + \mathbf{W}(\theta_1^i) A \odot \mathbf{W}(\theta_0^i) L_0,
\end{equation}
where $\mathbf{W}(\theta_l^i) \in \mathbb{R}^{n\times n}$ is a warping matrix according to the motion parameter $\theta_l^i$. The alpha blending mask $A$ is warped by the foreground motion $\theta_1^i$ since its appearance depends on the foreground object. Similarly to~\cite{xue2015computational}, for notational simplification, we redefine the clear foreground layer as $L_1 = (\mathbf{1} - A)\odot L_1$, and abbreviate the notation $\mathbf{W}(\theta_1^i)$ to $\mathbf{W}_1^i$. Then, the simplified equation is:
\begin{equation}
I^i = \mathbf{W}_1^i L_1 + \mathbf{W}_1^i A \odot \mathbf{W}_0^i L_0,
\label{eq_warp}
\end{equation}
Based on Eq.~\eqref{eq_warp}, we compare the previous layered blur model~\cite{wulff2014modeling} and our layered blur model in the following subsections.

\begin{figure*}
  \centering
  \includegraphics[width=\linewidth]{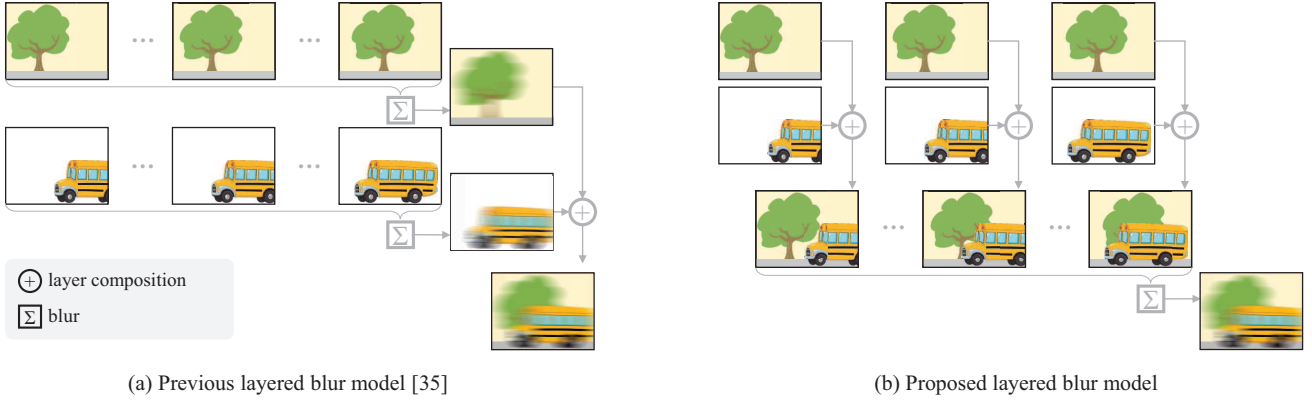}
  \caption{Comparison of two blur generative models. The order of layer composition and blur is changed in the previous model. The proposed model coincides with the actual blur generation process.}
  \label{fig_modelComparison}
\end{figure*}

\subsection{Previous Layered Blur Model}
Wulff and Black~\cite{wulff2014modeling} proposed a layered blur model to represent the mixture of the foreground and the background at occlusion boundaries. Their generative model addressed that blurred images consist of the composition of individually blurred foreground and background layers. To express this model, let $\mathbf{K}_l^i \in \mathbb{R}^{n\times n}$ denotes a blur matrix for each layer $l$ at frame $i$, which equals the average of warping matrices while the shutter is open, as:
\begin{equation}
\mathbf{K}_l^i = \frac{1}{T}\int_0^{T}\mathbf{W}(\theta_l^{i,t})dt,
\end{equation}
where $T$ is a exposure time, and $\theta_l^{i,t}$ is the motion parameter for intra-frame capture time $t$ (the elapsed time after the shutter of the frame $i$ is opened) from the reference frame (\ie $\theta_l^{i,0} = \theta_l^{i}$). Then, the blurred frame $B^i_{\text{prev}} \in \mathbb{R}^n$ based on~\cite{wulff2014modeling} is as follows:
\begin{equation}
B^i_{\text{prev}} = \mathbf{K}_1^i L_1 + \mathbf{K}_1^i A \odot \mathbf{K}_0^i L_0.
\label{eq_wulff}
\end{equation}

In this model, the blurred frame is the composition of individually blurred layers ($\mathbf{K}_1^i L_1$ and $\mathbf{K}_0^i L_0$). This equation is equivalent to the layered representation of motion-blurred video of~\cite{wulff2014modeling}, although the notation is different.

\subsection{Proposed Layered Blur Model}
Here we propose a new layered blur model. As shown in Figure~\ref{fig_counter}, the previous layered blur model does not reflect actual blur generation process in which a blurred image is generated by integrating intermediate images the camera sees during exposure. By applying this concept to the layered blur model, a blurred frame $B^i \in \mathbb{R}^n$ is newly represented as follows:

\begin{equation}
B^i = \frac{1}{T}\int_0^{T}\left( \mathbf{W}_1^{i,t} L_1 + \mathbf{W}_1^{i,t} A \odot \mathbf{W}_0^{i,t} L_0 \right) dt,
\label{eq_ours}
\end{equation}
where $\mathbf{W}_l^{i,t}$ is an abbreviated notation of $\mathbf{W}(\theta_l^{i,t})$.

In this model, the blurred frame is the integration of intermediate images, each of which is a composition of intermediate layers. This proposed model coincides with the actual blur generation process.

\begin{figure}
  \centering
  \includegraphics[width=\linewidth]{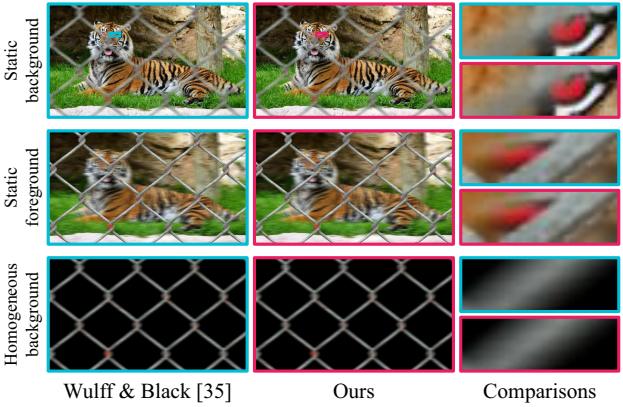}
  \caption{Three special cases that the model of Wulff and Black~\cite{wulff2014modeling} becomes equivalent to ours. Both blurred images are identical even at occlusion boundaries.}
  \label{fig_special}
\end{figure}

\subsection{Comparison of Layered Blur Models}
\label{section_comparison}
The main difference between the conventional layered blur model (Eq.~\eqref{eq_wulff}) and the proposed layered blur model (Eq.~\eqref{eq_ours}) is the order of a composition and a blurring of layers, as illustrated in Figure~\ref{fig_modelComparison}. While the conventional model composites two layers after blurring each layer (\ie integrating each intermediate layer), our model composites two layers first and then integrates the composited intermediate images.

Note that although Eq.~\eqref{eq_wulff} does not model actual physical two layered blurring process correctly, it becomes identical to Eq.~\eqref{eq_ours} in some special cases. We analyze and compare the two models, and show the conditions when the two models become equivalent, both analytically and empirically.

Due to the fact that \small
\begin{equation}
\left(\frac{1}{T}\int_0^{T} X(t) dt \right) \odot \left(\frac{1}{T}\int_0^{T} Y(t) dt\right) \neq \frac{1}{T}\int_0^{T} X(t) \odot Y(t) dt,
\label{eq_difference}
\end{equation}
\normalsize the blurred images generated by the two models in Eq.~\eqref{eq_wulff} and Eq.~\eqref{eq_ours} are different with each other as follows:
\begin{equation}
\begin{split}
B^i_\text{prev} &= \mathbf{K}_1^i L_1 + \mathbf{K}_1^i A \odot \mathbf{K}_0^i L_0 \\
                &\neq \mathbf{K}_1^i L_1 + \frac{1}{T}\int_0^{T}\left(\mathbf{W}_1^{i,t} A \odot \mathbf{W}_0^{i,t} L_0\right) dt = B^i.
\end{split}
\label{eq_comparison}
\end{equation}

Note, however, that the left and right formulas in Eq.~\eqref{eq_difference} become identical when $X(t)$ or $Y(t)$ is a constant with respect to $t$. And this leads to the following three cases that can make the two deblurring models in Eq.~\eqref{eq_comparison} become equivalent.
\begin{enumerate}
\item Background is static \footnotesize($\mathbf{W}_0^{i,t}$ is a constant \wrt $t$)\normalsize
\item Foreground is static \footnotesize($\mathbf{W}_1^{i,t}$ is a constant \wrt $t$)\normalsize
\item Background is homogeneous \footnotesize($\mathbf{W}_0^{i,t} L_0$ is a constant \wrt $t$)\normalsize
\end{enumerate}
Although homogeneous alpha map \footnotesize($\mathbf{W}_1^{i,t} A$ is a constant \wrt $t$)\normalsize~can also make two models become identical, it is impossible at occlusion boundaries.

Figure~\ref{fig_special} is the experimental comparison that shows the blurred images corresponding to these three situations. The two models give the same blurred images. Thus, the previous layered blur model is a good approximation of ours, but it may lead to artifacts at occlusion boundaries in general.

\section{Occlusion-Aware Video Deblurring}

In this section, we propose an occlusion-aware deblurring method based on the proposed layered blur model.

\subsection{Formulation}
Given a set of blurred frames including an occluding layer, we restore clear foreground, background, alpha blending mask, and object motions.

First we discretize the proposed model for deblurring. We divide the exposure time into $M$ samples uniformly; $\{\tau_k\}_{k=1}^M$ denote the sampled times such that $\tau_k = \frac{(k-1)}{M}T$. Then, our data term is defined as follows:
\begin{equation}
\sum_i\|\nabla B^i - \nabla\frac{1}{M}\sum_{k=1}^M\left( \mathbf{W}_1^{i,\tau_k} L_1 + \mathbf{W}_1^{i,\tau_k} A \odot \mathbf{W}_0^{i,\tau_k} L_0 \right) \|_2^2,
\label{eq_data}
\end{equation}
where $\nabla$ denotes a gradient operator, which is widely used in deblurring field to reduce ringing artifacts~\cite{cho2009fast,kim2015generalized,xu2013unnatural}. We use affine transformations for motion parameters and $\mathbf{W}_l^{i,\tau_k}$ is obtained by linearly interpolating the motions of adjacent frames $\theta_l^{i}$ and $\theta_l^{i+1}$~\cite{wulff2014modeling}.

Since the deblurring is a highly ill-posed problem, we add regularization terms to reduce the ambiguity. We enforce hyper-Laplacian priors~\cite{krishnan2009fast,krishnan2011blind,levin2007user} on the gradients of the foreground and the background images as follows:
\begin{equation}
\sum_l\|\nabla L_l \|_{0.8}^{0.8}.
\end{equation}

Also, since the alpha map is smoother than natural images, we use Laplacian prior on the gradient of $A$ as:
\begin{equation}
\|\nabla A \|_1.
\end{equation}

\begin{figure*}
  \centering
  \includegraphics[width=\linewidth]{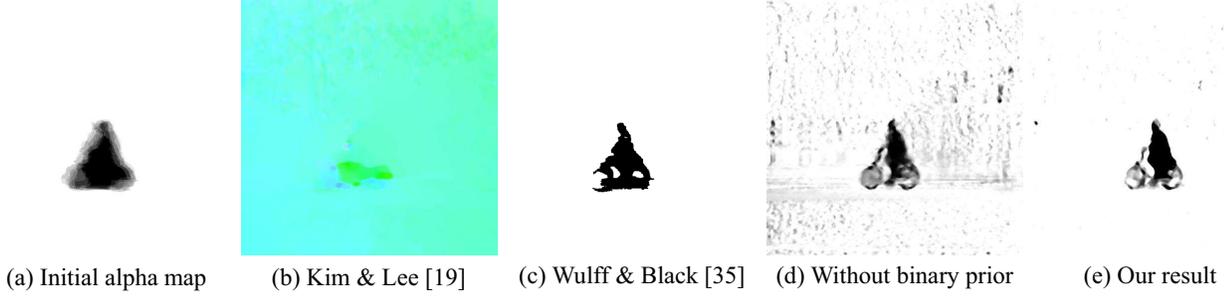}
  \caption{The contribution of the binary prior term for alpha blending mask. (a) In blurred video, the alpha map obtained through optical flow (Section~\ref{section_init}) does not accurately estimate the shape of the object. (b) State-of-the-art video deblurring~\cite{kim2015generalized} method do not correctly estimate the motion map. (c) The previous layered deblurring method~\cite{wulff2014modeling} restore the shape to some extent. (d) Even when there is no binary term (\ie $\lambda_3 = 0$), our method estimates a more detailed shape, but it is noisy. (e) Our method with the binary term provides a clear result.}
  \label{fig_binary}
\end{figure*}

Additionally, we enforce $A$ to be close to binary values by using following constraint:
\begin{equation}
A^T(\mathbf{1}-A).
\label{eq_binary}
\end{equation}
This term prevents the restored layers from being blurry transparent images. Figure~\ref{fig_binary} shows the contribution of the binary term Eq.~\eqref{eq_binary}.

The final objective function is as follows:
\begin{equation}
\begin{split}
\min_{L_0, L_1, A, \Theta}&\scriptstyle\lambda_1\sum_i\|\nabla B^i - \nabla\frac{1}{M}\sum_{k=1}^M\left( \mathbf{W}_1^{i,\tau_k} L_1 + \mathbf{W}_1^{i,\tau_k} A \odot \mathbf{W}_0^{i,\tau_k} L_0 \right) \|_2^2 \\
&\scriptstyle+ \sum_l\|\nabla L_l \|_{0.8}^{0.8} + \lambda_2\|\nabla A \|_1 + \lambda_3 A^T(\mathbf{1}-A),
\end{split}
\label{eq_objective}
\end{equation}
where $\lambda_1$, $\lambda_2$, and $\lambda_3$ are parameters that adjust the weight of each term, and $\Theta$ denotes a set of the motion parameters for each layer $l$ at each frame $i$ (\ie $\{\theta_l^i\}$). We restrict the intensities of $L_0$, $L_1$, and $A$ to be in the range [0, 1].

\subsection{Initialization}
~\label{section_init}
Since the problem is non-convex and easy to get stuck at a local minimum, it is important to start with good initial values. We use optical flow~\cite{xu2012motion} to initialize the motion parameters as done in~\cite{wulff2014modeling}. Although the optical flow does not handle blurred images correctly, it provides good initial values for the variables. Based on the initial optical flow, the two dominant affine transformations are estimated for each layer using RANSAC. Also, we initialize each layer as an average of aligned input frames using initial motion parameters for each layer~\cite{xue2015computational}. For the alpha blending mask, from the RANSAC result, we first specify the pixels that correspond to the background of each frame to the intermediate masks. Then, we initialize alpha blending mask as an average of the aligned intermediate masks using the motion parameters for the background.

\subsection{Optimization}
To optimize the non-convex objective function in Eq.~\eqref{eq_objective}, we divide the original problem into three sub-problems and use alternating optimization techniques~\cite{cho2009fast,kim2013dynamic,wulff2014modeling} that iteratively estimate each unknown while other unknowns are fixed. In this section, we reorganize each sub-problem as a traditional deconvolution formula and describe how to solve it. Notice that we estimate non-simplified $\{L_1, L_0, A\}$ (\ie $I = (1-A) \odot L_1 + A \odot L_0$) although we used the simplified formula (\ie $I = L_1 + A \odot L_0$) in the previous section for notational simplicity.

\paragraph{Latent Image Estimation}
In this step, we restore clear foreground and background layers while alpha blending mask and motion parameters are fixed. To make this sub-problem be in a simpler formula, we concatenate some variables. Let $L \in \mathbb{R}^{2n}$ denotes the row concatenation of $L_0$ and $L_1$ (\ie \small$L = \begin{bmatrix} L_0 \\ L_1 \end{bmatrix}$\normalsize), and $\mathbf{K}_L^i \in \mathbb{R}^{n \times 2n}$ denotes the column concatenation of $\mathbf{K}_{L_0}^i \in \mathbb{R}^{n \times n}$ and $\mathbf{K}_{L_1}^i \in \mathbb{R}^{n \times n}$ (\ie \small$\mathbf{K}_L^i = \begin{bmatrix} \mathbf{K}_{L_0}^i & \mathbf{K}_{L_1}^i \end{bmatrix}$\normalsize) such that
\begin{equation}
\begin{split}
\mathbf{K}_{L_0}^i &= \frac{1}{M}\sum_{k=1}^M\text{diag}\left(\mathbf{W}_1^{i,\tau_k}A\right)\mathbf{W}_0^{i,\tau_k},\\
\mathbf{K}_{L_1}^i &= \frac{1}{M}\sum_{k=1}^M\mathbf{W}_1^{i,\tau_k}\text{diag}(1-A),
\end{split}
\end{equation}
where $\text{diag}(\cdot)$ denotes a diagonal matrix formed from its vector argument. Since $X \odot Y$ can be represented as $\text{diag}(X)Y$, $\mathbf{K}_L^i L$ is equivalent to our layered blur model.

Then, for multi-frames, let $B \in \mathbb{R}^{Nn}$ denotes the row concatenation of $\{B^i\}$, and $\mathbf{K}_L \in \mathbb{R}^{Nn \times 2n}$ denotes the row concatenation of $\{\mathbf{K}_L^i\}$ ($N$ is the number of observed frames). The sub-problem for latent image estimation can be expressed as follows:
\begin{equation}
\min_L \lambda_1\|\nabla B - \nabla \mathbf{K}_L L \|_2^2 + \|\nabla L \|_{0.8}^{0.8}.
\label{eq_L}
\end{equation}
This optimization problem is same as the traditional deconvolution problem with hyper-Laplacian prior~\cite{krishnan2009fast}. We optimize Eq.~\eqref{eq_L} using a conjugate gradient method and a lookup table in the same way as~\cite{krishnan2009fast}.

\paragraph{Alpha Blending Mask Estimation}
In this step, we restore clear alpha blending mask while layer appearances and motion parameters are fixed. Similarly to the latent image estimation step, we define $B_{A}^i \in \mathbb{R}^{n}$ and $\mathbf{K}_{A}^i \in \mathbb{R}^{n \times n}$ as follows:
\begin{equation}
\begin{gathered}
B_{A}^i = B^i - \frac{1}{M}\sum_{k=1}^M\mathbf{W}_1^{i,\tau_k}L_1,\\
\mathbf{K}_{A}^i = \frac{1}{M}\sum_{k=1}^M\left(\text{diag}\left(\mathbf{W}_0^{i,\tau_k}L_0\right)\mathbf{W}_1^{i,\tau_k} - \mathbf{W}_1^{i,\tau_k}\text{diag}(L_1)\right).
\end{gathered}
\end{equation}
Also, let $B_{A} \in \mathbb{R}^{Nn}$ denotes the row concatenation of $\{B_{A}^i\}$, and $\mathbf{K}_{A} \in \mathbb{R}^{Nn \times n}$ denotes the row concatenation of $\{\mathbf{K}_{A}^i\}$. Then, the sub-problem for alpha map estimation can be expressed as follows:
\begin{equation}
\min_A \lambda_1\|\nabla B_A - \nabla \mathbf{K}_{A} A\|_2^2 + \lambda_2\|\nabla A \|_1 + \lambda_3 A^T(\mathbf{1}-A)
\label{eq_A}
\end{equation}
This sub-problem is also same as a traditional $L1$ deconvolution problem except the last term. We optimize Eq.~\eqref{eq_A} using a primal-dual optimization method~\cite{chambolle2011first} as follows:
\begin{equation}\begin{cases}
D^{m+1} = \frac{D^m + \sigma_D\nabla \overline{A}^m}{\max(\mathbf{1}, |D^m + \sigma_D\nabla \overline{A}^m|)}\\
\\
\begin{split}
A^{m+1} = \argmin_A& \frac{\lambda_2\| A - (A^m - \sigma_A\nabla^T D^{m+1}) \|_2^2}{2\tau} +\\
&\lambda_1\|\nabla B_A - \nabla \mathbf{K}_{A} A\|_2^2 + \lambda_3 A^T(\mathbf{1}-A)
\end{split}\\
\\
\overline{D}^{m+1} = 2D^{m+1} - \overline{D}^{m},
\end{cases}
\label{eq_A_pd}
\end{equation}
where $m$ denotes the iteration number, $D \in \mathbb{R}^{n}$ denotes the dual variable, and $\sigma_D = 10$ and $\sigma_A = 0.0125$ are the parameters for each update step. We apply the conjugate gradient method to optimize $A^{m+1}$.

\paragraph{Motion Parameter Estimation}
In this step, we estimate motion parameters while layer appearances and alpha blending mask are fixed. If we focus on the terms dependent on $\Theta$ in Eq.~\eqref{eq_objective}:
\begin{equation}
\min_{\Theta}\scriptstyle\sum_i\|\nabla B^i - \nabla\frac{1}{M}\sum_{k=1}^M\left( \mathbf{W}_1^{i,\tau_k} L_1 + \mathbf{W}_1^{i,\tau_k} A \odot \mathbf{W}_0^{i,\tau_k} L_0 \right) \|_2^2,
\label{eq_theta}
\end{equation}
We solve this equation using the Nelder-Mead simplex method~\cite{lagarias1998convergence} that can be implemented simply by a Matlab built-in function ``fminsearch".

\renewcommand{\algorithmicrequire}{\textbf{Input:}}
\renewcommand{\algorithmicensure}{\textbf{Output:}}
\begin{algorithm}
\caption{Overview of the proposed deblurring method}
\begin{algorithmic}[1]
\REQUIRE A set of blurred frames $\{B^i\}$
\ENSURE Latent layers $L$, mask $A$, and motions $\Theta$.
\STATE Initialize $L$, $A$, $\Theta$ using optical flow.
\STATE Build image pyramid.
\STATE \textbf{for} iteration = 1:3
\STATE \quad Solve for $L$ (Eq.~\eqref{eq_L}).
\STATE \quad Solve for $A$ (Eq.~\eqref{eq_A_pd}).
\STATE \quad Solve for $\Theta$ (Eq.~\eqref{eq_theta}).
\STATE \textbf{end}
\STATE Propagate variables to the next pyramid level if exists.
\STATE Repeat steps 3-8 from coarse to fine pyramid level.
\end{algorithmic}
\label{algorithm_overview}
\end{algorithm}

\subsection{Implementation Details}
To accelerate the algorithm, we optimize the objective function based on coarse-to-fine approach where the scale factor of image pyramid is 0.8. Also, the parameters used in the optimization is fixed for all experiments as $\lambda_1 = \frac{2500}{N}$ and $\lambda_3 = \frac{\lambda}{20000}$,  where $N$ denotes the number of frames. $\lambda_2$ was adjusted to a value between $0.05\lambda$ and $0.06\lambda$ depending on the shape of the occluding object. Camera duty cycle, which is the ratio of an exposure time $T$ to a frame interval, is given from the camera setting (0.5). The overview of our algorithm is summarized in Algorithm~\ref{algorithm_overview}.


\begin{figure*}
  \centering
  \includegraphics[width=\linewidth]{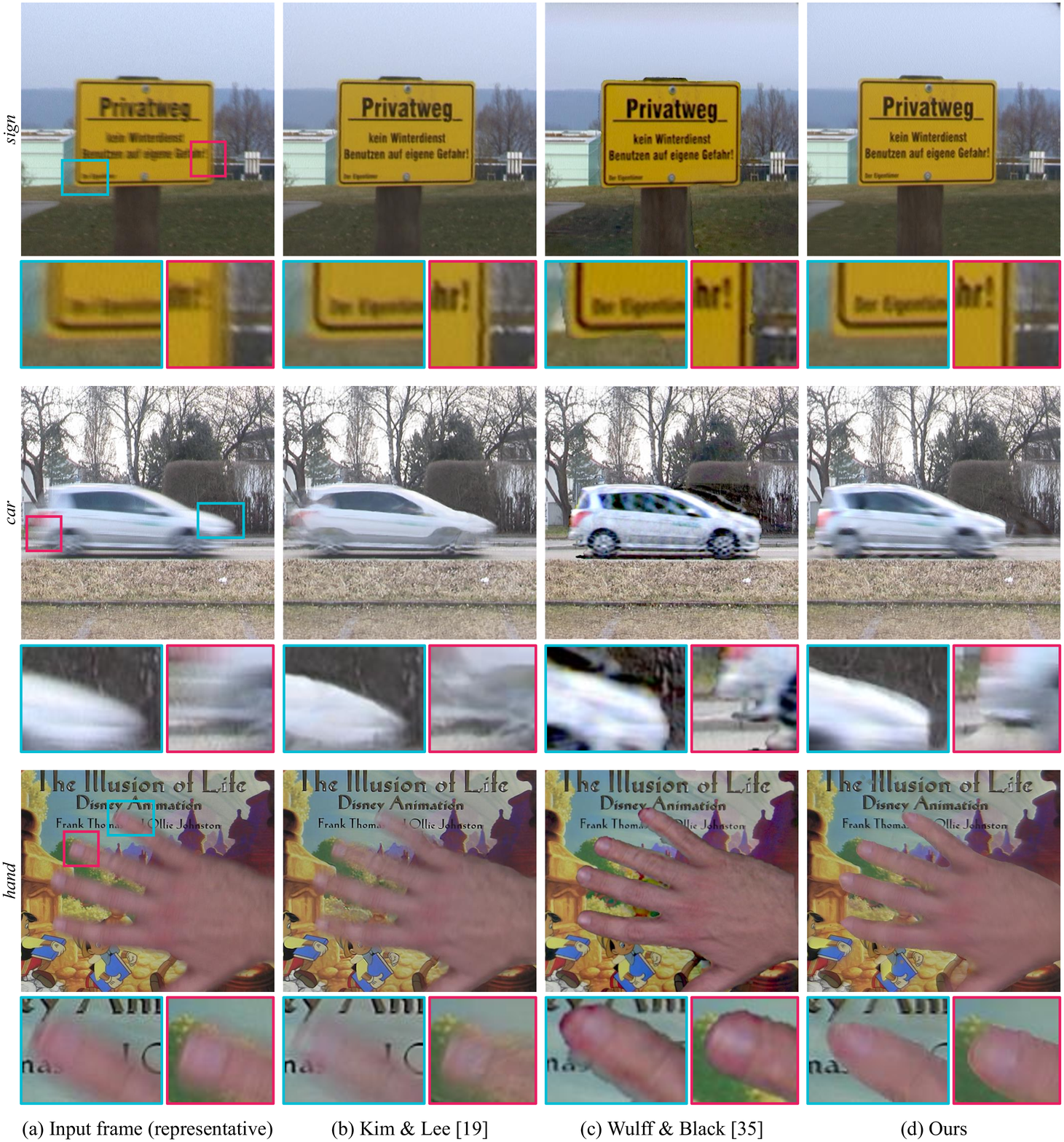}
  \caption{Comparison with recent methods. The previous method~\cite{kim2015generalized,wulff2014modeling} causes segmentation errors and consequent artifacts at occlusion boundaries, but our method yields clear results.}
  \label{fig_results}
\end{figure*}


\begin{figure*}
  \centering
  \includegraphics[width=\linewidth]{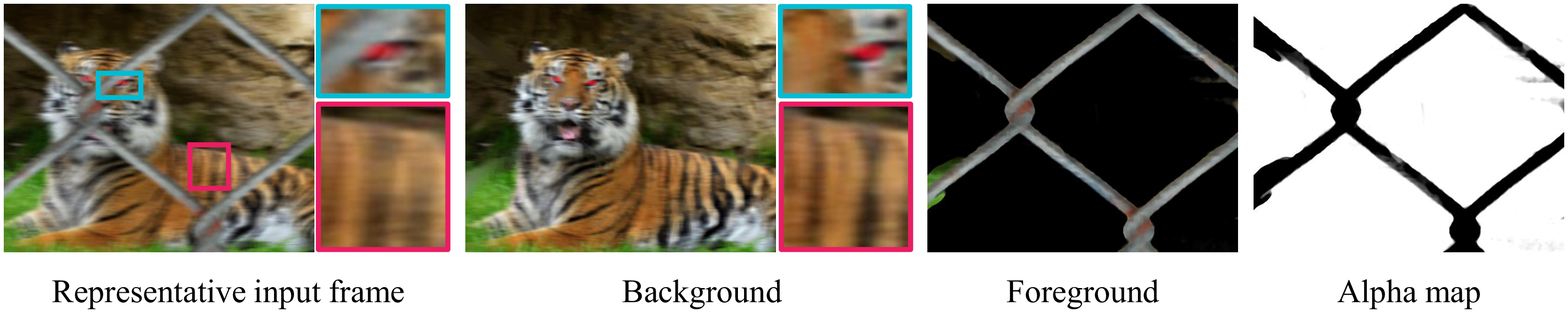}
  \caption{Occlusion removal result. Our deblurring method estimates clean foreground and background separately, so it is possible to achieve the effects of layer separations~\cite{xue2015computational} as well. When a tiger is blurred while being occluded by a fence, we can separate the tiger in the background and foreground fence, and restore clear images of them at the same time.}
  \label{fig_result_layer}
\end{figure*}

\section{Experimental Results}
We compare our deblurring results with those of the state-of-the-art video deblurring method~\cite{kim2015generalized} and the previous layered deblurring method~\cite{wulff2014modeling}. Since the source code of the previous method~\cite{wulff2014modeling} is not available, we focus on comparing our results with published results from the previous methods. Please see our supplementary material for more results.

The image sequences consist of 5 to 10 images. We processed the sequences on a desktop computer with Intel i7-6700k CPU and 64GB memory. It took 10 minutes per image to process 640x480 images with a non-optimized Matlab implementation.

Figure~\ref{fig_teaser} and Figure~\ref{fig_binary} show the deblurring results for images with severe occlusions. Since the detailed structure of the bicycle is mixed with the background by the occlusion, generalized deblurring method~\cite{kim2015generalized} that does not consider the occlusion attempts to restore the background mainly. Also, since both the foreground and background are blurred, the previous layered deblurring method~\cite{wulff2014modeling} do not restore the details on the face and the bicycle properly. On the other hand, the proposed method restores the detailed structure of the human face and bicycle handle by using the carefully designed model.

Figure~\ref{fig_results} shows the comparision with recent methods for deblurring results. The "sign" sequence and "car" sequence correspond to the situations where both foreground and background are moving. In the result of the previous method~\cite{kim2015generalized,wulff2014modeling}, we can see the artifacts caused by segmentation errors at boundaries. Our result shows better performance at occlusion boundaries. Also, the "hand" sequence belongs to a situation where the previous model and our model are the same because the background is static. However, even in this situation, we can see that our method produces better results because it uses effective regularization and optimization.

Additionally, our method can also achieve the effects of layer separations~\cite{xue2015computational}. It can remove occluding objects even when the images are blurry. Figure~\ref{fig_result_layer} shows not only that the fence occluding the tiger is removed, but also that the image is clearly restored. This image sequence was created by a physics-based renderer~\cite{jakob2010mitsuba}.

\section{Discussion}
In this section, we analyze the blur kernel at object boundaries, which shows the distinctive characteristics of the blur kernel that cannot be captured by conventional blur models. In addition, we discuss the limitations and future works of the proposed method.

\subsection{Blur Kernel at Occlusion Boundaries}
\begin{figure}
  \centering
  \includegraphics[width=\linewidth]{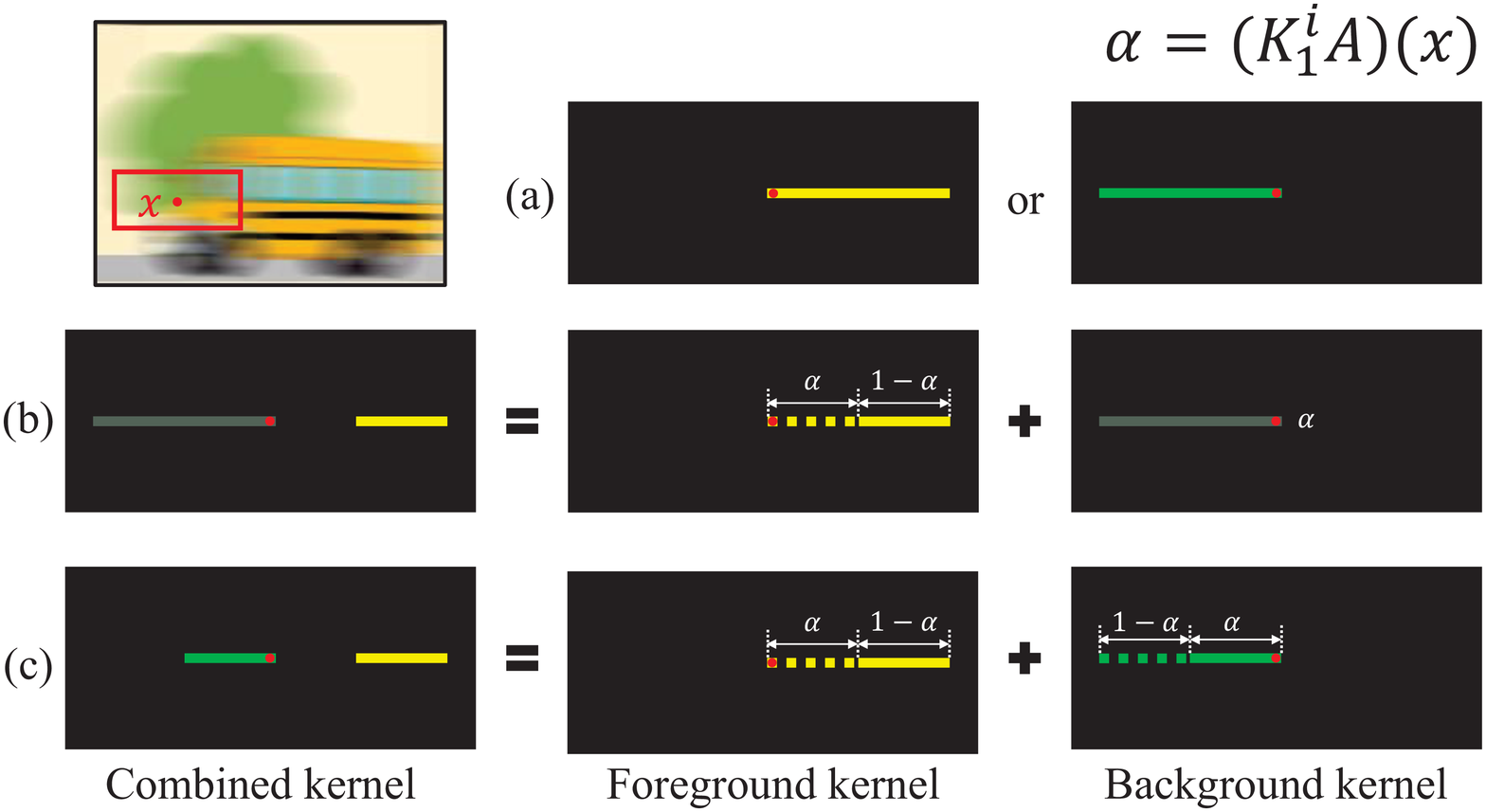}
  \caption{Illustration of blur kernels at boundaries according to each model. (a) Early models. (b) Wulff and Black~\cite{wulff2014modeling}. (c) Proposed model. The proposed model correctly models the occlusion event with the length of the blur kernel instead of its intensity~\cite{wulff2014modeling} according to the blurred mask value $\alpha$. }
  \label{fig_discussion}
\end{figure}

Visualizing the proposed blur model as a traditional blur kernel gives us the interesting result at layer boundaries. Figure~\ref{fig_discussion} illustrates the blur kernel of each model at boundaries where the foreground is moving to the left-side and the background is moving to the right-side.

Early works that handle abruptly-varying blur find the kernel either of the foreground or of the background~\cite{chakrabarti2010analyzing,cho2007removing,dai2009removing,kim2013dynamic,lee2013dense,paramanand2013non} such as Figure~\ref{fig_discussion}(a), or find an ambiguous kernel between them~\cite{kim2014segmentation,kim2015generalized}.

In the model of Wulff and Black~\cite{wulff2014modeling}, the pixel at boundaries is blurred by both foreground and background kernels while each kernel is truncated or diminished in intensity compared to the kernel without occlusion. The foreground kernel is shortened in length to a factor of $1-\alpha$ and the background kernel is reduced in intensity to a proportion of $\alpha$ as shown in Figure~\ref{fig_discussion}(b), where $\alpha$ is the blurred mask value of the corresponding pixel.

In the proposed model, the foreground kernel experiences truncation equal to that of~\cite{wulff2014modeling}, but the background kernel is \emph{truncated} instead of being \emph{weakened} to a factor of $\alpha$ as shown in Figure~\ref{fig_discussion}(c). In representing the occlusion of a background by a foreground object, the proposed model correctly models the occlusion event with \emph{the length of the blur kernel}.

Thus, the blur kernel of the foreground and the background can be \emph{separate} or \emph{overlapped} each other according to the relative velocity of the layers. These distinctive kernel characteristics cannot be captured by conventional blur models. 

\subsection{Limitation and Future work}
In this study, several assumptions are made for our deblurring method. We assumed that the camera duty cycle is given for every frame, and the object motion is smooth in a frame. Since we fix the camera setting, and the exposure time is short enough in videos, these assumptions can be justified to some extent. For the videos without exposure information, the estimation of duty cycle~\cite{li2010generating} should be combined to our method.

Also, we parameterized the object motion as an affine motion, which causes a limitation in dealing with general object motions. Although a projective motion can be applied to our model, a further consideration is required for additional problems such as occlusions in a layer or brightness constancy. Combining our method with a non-parametrical motion deblurring method~\cite{kim2014segmentation,kim2015generalized} is one of our future directions. In addition, our method assumes the scene with two layers currently. Expanding this to multi-layer requires additional consideration of the occlusions involving multi-layer and the depth order of the layers. Solving this problem is our another future direction.


\section{Conclusion}
In this paper, we proposed occlusion-aware video deblurring based on a new layered blur model, allowing us an accurate restoration of object boundaries. We addressed the limitation of the conventional layered blur model theoretically and experimentally, and enhanced the model by changing the order of layer composition and blur, so that it follows the actual blur generation process. Based on this model, the proposed occlusion-aware deblurring method obtains more accurate latent image, object motion, and segmentation mask. Also, we analyzed that our model exactly extracts the contribution of occlusion from the original kernel, helping the capture of the property to overlap or separate the foreground and background kernels at boundaries. Experimental results on synthetic and real blurred videos demonstrate the outstanding performance of the proposed method.


\end{document}